\documentclass{article}

\usepackage{arxiv}

\usepackage[utf8]{inputenc} 
\usepackage[T1]{fontenc}    
\usepackage{hyperref}       
\usepackage{url}            
\usepackage{booktabs}       
\usepackage{amsfonts}       
\usepackage{nicefrac}       
\usepackage{microtype}      
\usepackage{lipsum}		

\usepackage{multirow}
\usepackage{enumitem}
\usepackage{amsmath}
\usepackage{comment}
\usepackage{fancyvrb}

\usepackage{graphicx}
\usepackage{natbib}
\usepackage{doi}
\usepackage{listings}
\usepackage{xcolor}
\title{Evidence‑Driven Reasoning for Industrial Maintenance Using Heterogeneous Data}

\author{
\parbox{0.75\textwidth}{\centering
Fearghal O'Donncha$^{1}$\thanks{Corresponding author: feardonn@ie.ibm.com},
Nianjun Zhou$^{1}$,
Natalia Martinez$^{2}$,
James T Rayfield$^{2}$,
Fenno F. Heath III$^{2}$,
Abigail Langbridge$^{3}$,
Roman Vaculin$^{2}$ }
}

\date{
$^{1}$IBM Research Europe, Ireland 
$^{2}$IBM Research Yorktown, USA 
$^{3}$Imperial College London, UK
}

\renewcommand{\shorttitle}{Evidence-Driven Reasoning for Industrial Maintenance}

\hypersetup{
pdftitle={A template for the arxiv style},
pdfsubject={q-bio.NC, q-bio.QM},
pdfauthor={David S.~Hippocampus, Elias D.~Striatum},
pdfkeywords={First keyword, Second keyword, More},
}

\begin{document}
\maketitle

\begin{abstract}
Industrial maintenance platforms contain rich but fragmented evidence, including free-text work orders, heterogeneous operational sensors or indicators, and structured failure knowledge. These sources are often analyzed in isolation, producing alerts or forecasts that do not support conditional decision-making: given this asset history and behavior, what is happening and what action is warranted?

We present Condition Insight Agent, a deployed decision-support framework that integrates maintenance language, behavioral abstractions of operational data, and engineering failure semantics to produce evidence-grounded explanations and advisory actions. The system constrains reasoning through deterministic evidence construction and structured failure knowledge, and applies a rule-based verification loop to suppress unsupported conclusions.

Case studies from production CMMS deployments show that this verification-first design operates reliably under heterogeneous and incomplete data while preserving human oversight. Our results demonstrate how constrained LLM-based reasoning can function as a governed decision-support layer for industrial maintenance.
\end{abstract}


\section{Introduction}
\label{sec:intro}



Enterprise asset-intensive organizations rely on condition-based maintenance to reduce downtime and manage operational risk. Computerized Maintenance Management Systems (CMMS) aggregate health scores, alerts, work orders, and operational indicators, but rarely support the conditional reasoning required for maintenance decisions.

In practice, decisions depend on reasoning across multiple forms of evidence: given a change in behavior, which failure hypotheses are plausible, which align with maintenance history, and what action is appropriate? Existing systems fall short because evidence remains fragmented across meters, work orders, and engineering knowledge, while relationships between operational signals and asset models are often uncertain. Practitioners must manually reconcile these sources under time pressure reliant on expert judgment.

This challenge is compounded by industrial data realities: operational indicators originate from heterogeneous IoT and SCADA (Supervisory Control and Data Acquisition) platforms with inconsistent naming and limited shared ontology; maintenance history is largely unstructured text; and engineering artifacts such as Failure Modes and Effects Analysis (FMEA) are rarely integrated into data-driven workflows. Supporting maintenance reasoning therefore requires controlled reasoning over heterogeneous, partially structured, and semantically uncertain evidence. Large language models (LLMs) offer flexible reasoning, but naively applying generative models in reliability-critical settings introduces risks. Unconstrained agents may hallucinate explanations or produce fluent but unsupported recommendations. In industrial maintenance, trust depends on traceability, engineering consistency, and explicit linkage between claims and evidence.

We present \emph{Condition Insight}, a deployed reasoning framework that combines deterministic evidence construction with constrained LLM-based synthesis. Meter histories are abstracted into behavioral summaries, work orders into maintenance patterns, and FMEA-derived semantics bound the space of admissible explanations. The system is integrated into enterprise CMMS workflows and evaluated across large asset portfolios.

Our solution and contributions for this paper are:
\begin{itemize}
    \item We define Conditional Insight as a deployed decision-support task connecting operational indicators, maintenance history, and engineering failure knowledge into evidence-grounded explanations.
    \item We introduce a trajectory-controlled architecture separating deterministic evidence extraction from constrained LLM synthesis governed by failure semantics.
    \item We introduce a deterministic verification loop that cross-checks LLM outputs against operational rules and structured evidence, ensuring auditable decision elements.
    \item We demonstrate how this verification-first design improves traceability and hallucination resistance compared to unconstrained LLM pipelines.
    \item We report deployment lessons from enterprise CMMS environments with heterogeneous IoT data and sparse ontology support.
\end{itemize}

Beyond this application, our results illustrate how constrained, evidence-grounded LLM reasoning can be operationalized in reliability-critical environments where explanations must remain auditable.

\section{Related Work}
\label{sec:related-work}

\textbf{Data-Driven and Knowledge-Driven Maintenance.}
Condition-based and predictive maintenance research has focused on anomaly detection, fault diagnosis, and remaining useful life estimation from sensor data \cite{carvalho2019systematic, lei2018machinery, teixeira2020condition}. These approaches typically produce alerts or forecasts rather than integrated explanations that connect observed behavior to maintenance history and engineering failure semantics. In parallel, knowledge-driven methods employ rule-based systems, ontologies, digital twins, and FMEA to encode asset structure and causal failure knowledge \cite{lu2020digital, langbridge2023causal}. While interpretable, such approaches often depend on comprehensive ontologies and consistent data alignment, which are difficult to sustain in enterprise environments with heterogeneous assets and evolving data schemas \cite{moosavi2024explainable, ahmed2022artificial}.

\textbf{LLMs, Agents, and Constrained Generation.}
Large language models (LLMs) enable multi-step reasoning through retrieval-augmented generation and tool-augmented agents \cite{brown2020language, lewis2020retrieval, yao2022react}. These capabilities have motivated applications in industrial operations, including agentic anomaly detection and asset-oriented benchmarks \cite{patel2025assetopsbench, timms2024agentic, hosseini2025role}. However, many agentic systems prioritize autonomy over constraint, which poses risks in reliability-critical settings \cite{acharya2025agentic, ferrag2025llm}. Recent work on reliable generation introduces verification loops, self-refinement, human-in-the-loop evaluation, and LLM-based judging to improve grounding and reduce hallucination \cite{madaan2023self, zheng2023judging, wang2020human, langer2021future}. Our work extends this line by demonstrating a deployed reasoning framework in which deterministic evidence construction and rule-based governance explicitly bound LLM synthesis under real-world maintenance constraints.

\section{Problem Setting and Task}
\label{sec:problem-data}
We study conditional reasoning for maintenance decision support in an enterprise CMMS deployment spanning multiple asset classes. Rather than prediction, the objective is decision-ready interpretation of asset condition under heterogeneous and imperfect evidence.

\noindent
\textbf{Condition Insight Task}

Given maintenance evidence (historical work orders, asset-level operational indicators, and structured failure knowledge) the task is to generate an interpretable explanation of an asset’s current or emerging condition together with appropriate maintenance recommendations. The output is an evidence-grounded narrative linking observed behavior to maintenance history and plausible failure mechanisms.

\noindent
\textbf{Evidence Sources}
\noindent
The system operates over three complementary but weakly aligned evidence types:
\begin{itemize}
    \item \textbf{Work orders} document past symptoms, diagnoses, and interventions in largely unstructured text, capturing experiential maintenance knowledge.
    \item \textbf{Operational indicators (“meters”)} are low-frequency, curated measurements (e.g., runtime, periodic readings, discrete states) whose meaning arises from temporal patterns and whose semantics vary across cumulative, incremental, and categorical forms.
    \item \textbf{Failure knowledge} is represented using Failure Modes and Effects Analysis (FMEA), linking asset components to plausible failure mechanisms and influencing factors that constrain admissible explanations.
\end{itemize}

\noindent
\textbf{Integration Constraints}
\noindent
In practice, these sources are not natively aligned. Operational indicators originate from heterogeneous IoT and SCADA systems with inconsistent naming and limited shared ontology, and mappings between telemetry points and CMMS assets or components are often partial or ambiguous. Consequently, schema-level integration is unreliable and rule-based reasoning over raw signals is brittle. Conditional reasoning must therefore operate on abstracted behavioral evidence rather than authoritative semantic mappings while tolerating incomplete context and uncertain data relationships.

\section{Condition Insight Framework}
\label{sec:system}

The Condition Insight system separates deterministic evidence construction from constrained LLM reasoning. This separation reflects a key deployment requirement: explanations must remain traceable to operational evidence and consistent with engineering knowledge, even when underlying data is heterogeneous or incomplete.

Figure~\ref{fig:system_overview} illustrates the pipeline. Heterogeneous asset data (including work orders, operational indicators, and failure knowledge) are first transformed into a structured evidence representation. A domain LLM agent then performs constrained synthesis over this evidence, followed by a verification stage that governs key decision elements.

\subsection{Overview of Reasoning Pipeline}

The system operates in two stages:

\begin{itemize}
    \item \textbf{Analytics and Evidence Construction} produces a structured evidence packet (\texttt{asset\_facts}) summarizing asset behavior, maintenance history, and failure semantics.

    \item \textbf{Constrained LLM Reasoning} generates condition insights and recommendations using this evidence, while deterministic checks verify adherence to operational criteria.
\end{itemize}

This separation decouples statistical signal extraction from semantic interpretation, allowing the LLM to reason over curated evidence rather than raw operational streams.

\subsection{Deterministic Evidence Construction}

This stage converts heterogeneous maintenance data into interpretable behavioral evidence.

\noindent
\textbf{Work-order abstraction} extracts recurring symptoms, interventions, and outcomes from historical records, forming patterns of prior maintenance experience.

\noindent
\textbf{Meter abstraction} transforms low-frequency operational indicators into behavioral summaries rather than raw readings. Depending on meter semantics, this includes trend detection, change points, reset identification, drift characterization, and anomaly patterns. These summaries capture usage and condition evolution while suppressing noise. Additional details on meter abstraction is provided in Appendix \ref{sec:appendix_meter_abstraction}

\noindent 
\textbf{Failure knowledge alignment} links work-order evidence to failure mechanisms through a distribution-aware semantic matching process based on an Unbalanced Optimal Transport (UOT) \cite{chizat2018scaling} approach. Work-order descriptions $\mathbf{w}$ and failure mechanisms and degradation modes $\mathbf{m}$ are embedded into a shared vector space, and the optimal matching between the \textit{distributions} of $\mathbf{w}$ and $\mathbf{m}$ is computed as follows: 
\begin{equation}
\begin{aligned}
    \min_{\mathbf{T} : \mathbf{T} \geq 0} \mathbf{C} \cdot \mathbf{T} - \varepsilon H(\mathbf{T}) & + \rho_1 \mathrm{KL}(\mathbf{T} \mathbf{1}, \mathbf{w})\\
    &+ \rho_2 \mathrm{KL}(\mathbf{T}^T \mathbf{1}, \mathbf{m}),
\end{aligned}
\label{eq:uot}
\end{equation}
where $\mathbf{C}$ is a squared Euclidean cost on the latent space, $H(\cdot)$ is the Shannon entropy, $\mathrm{KL}(\cdot)$ are Kullback-Leibler divergences, and $\mathbf{T}$ is the optimal matching which defines the semantic correspondence. This process constructs a structured hypothesis space connecting observed behavior to engineering-grounded failure semantics.

All processing in this stage is deterministic, producing auditable signals that serve as inputs to downstream reasoning.

\begin{figure}[t]
  \centering
  \includegraphics[width=\columnwidth]{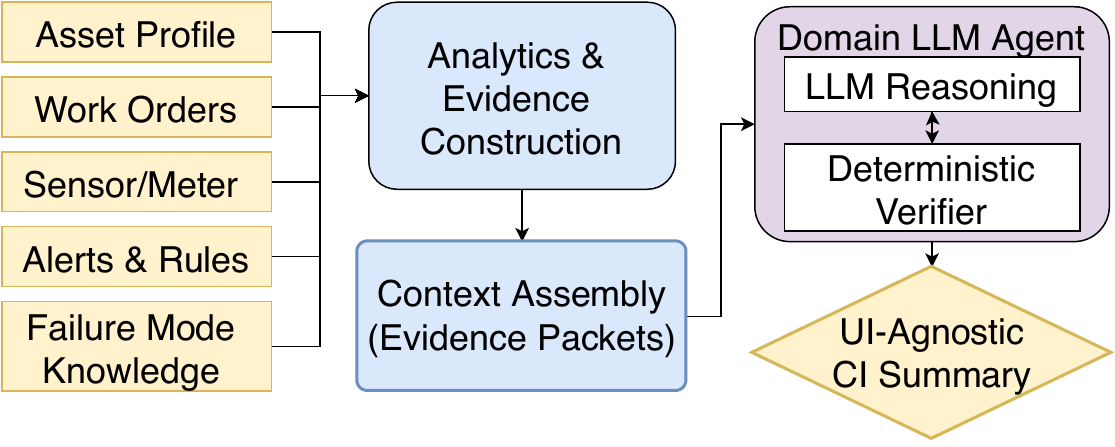}
  \caption{End‑to‑end pipeline of the Condition Insight System. Heterogeneous asset data and failure‑mode knowledge are consolidated into structured evidence packets, which are processed by a domain LLM agent with verifier‑based checking to produce a UI‑agnostic condition insight summary.
  }
  \label{fig:system_overview}
\end{figure}

\subsection{Domain LLM Agent and Deterministic Verification}

The Domain LLM Agent consumes the structured evidence packet and produces a Condition Insight Summary consisting of evidence-backed insights and recommendations. The LLM operates under domain-specific constraints that restrict explanations to those supported by evidence and consistent with failure semantics. For ablation, we compare this Constrained configuration with a Naive prompt that requests a general condition summary over the same structured evidence packet but does not encode explicit rule-aligned condition criteria or failure-mode constraints. The deterministic verification loop remains active in both settings, ensuring identical governance checks while isolating the effect of prompt-level constraints. Full details on the prompt and strcutured evidence packet is given in Appendix \ref{sec:appendix-schema}

In production deployment, key decision elements are governed through a deterministic verification loop. The LLM is prompted to assign an overall condition category (e.g., \emph{Normal}, \emph{Needs Attention}, \emph{Not Enough Data}) based on maintenance signals such as open or delayed work orders, emergency interventions, alerts, and meter anomalies. These same criteria are implemented as explicit rules in a parallel deterministic pipeline. The system compares the LLM-generated condition with the rule-based outcome, using discrepancies to identify reasoning errors and refine prompts.

This generation–verification separation ensures that high-level categorization remains governed by explicit operational criteria, while the LLM contributes explanatory synthesis over heterogeneous evidence.

\section{Case Studies}
\label{sec:casestudies}

We evaluated Condition Insight in two deployment settings to assess whether the verification-first reasoning architecture operates reliably under production data constraints. These studies examine evidence grounding, behavior under uncertainty, and interaction with practitioner workflows rather than model capability in isolation.

\subsection{Case Study A: Product Integration}

The system was deployed as an early-access decision-support capability within an enterprise CMMS, operating solely on production data available to practitioners (work orders, maintenance status, asset metadata, and condition indicators). As typical in such environments, evidence was incomplete and uneven.

The system produced structured condition summaries linking operational disruptions, maintenance balance, execution delays, and diagnostic coverage gaps to explicit evidence. All outputs passed through deterministic verification checks, and unsupported conclusions were revised or suppressed. Recommendations remained advisory and aligned with existing workflows; no automated actions were triggered. The system maintained consistent insight structure across asset types while surfacing uncertainty and data limitations as explicit outputs.

\begin{figure}[t]
  \centering
  \includegraphics[width=\columnwidth]{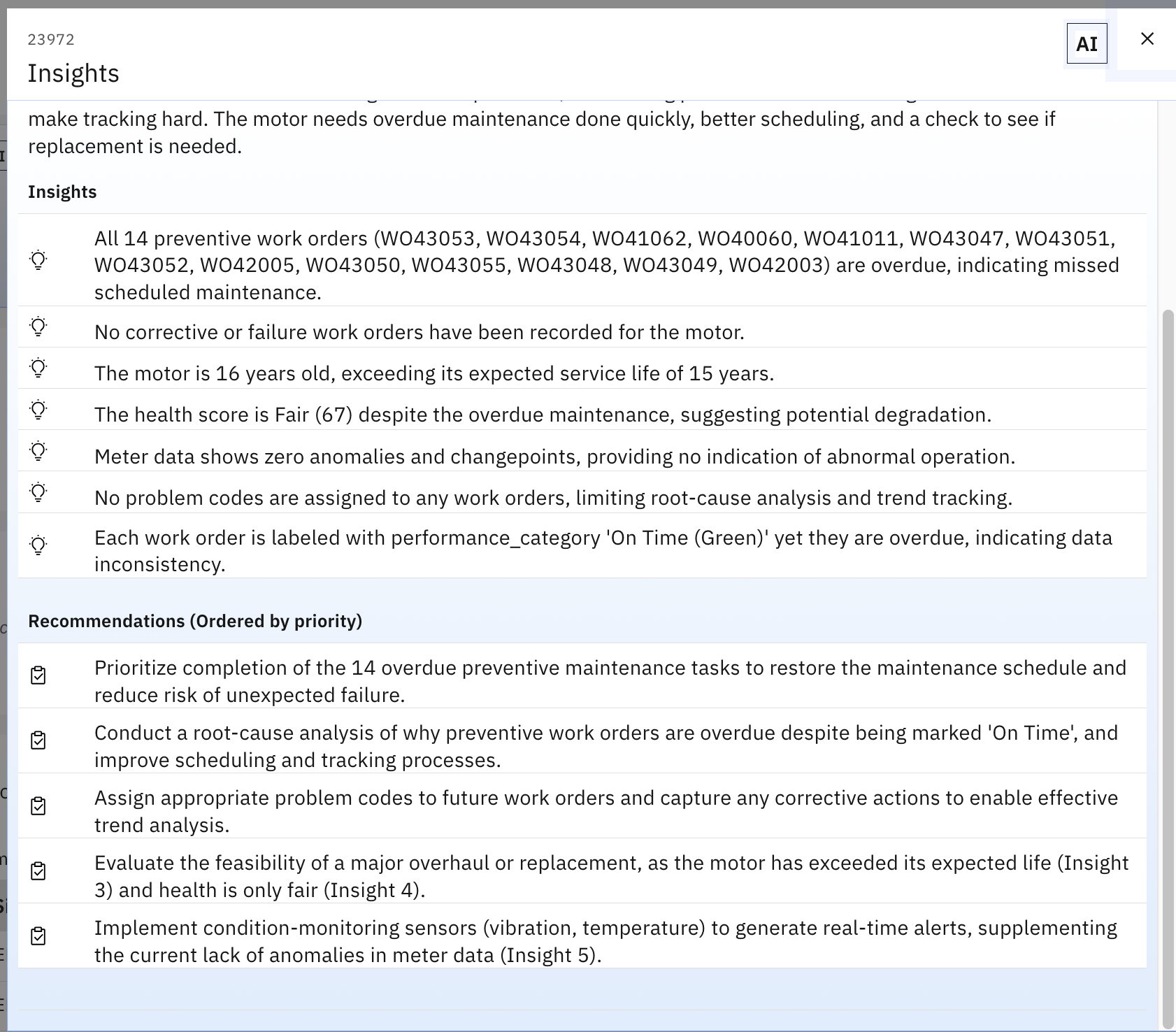}
  \caption{Example condition insight generated by the proposed framework using historical work orders of the last 200 days for a motor asset. 
  The output separates evidence-based observations (top) from prioritized, actionable recommendations (bottom), 
  synthesizing maintenance history, asset metadata, health indicators, and data consistency checks into a decision-oriented summary.}
  \label{fig:example_wo_insight}
\end{figure}

Figure~\ref{fig:example_wo_insight} shows an example output. The system produces a structured, UI-agnostic summary consisting of evidence-based insights and prioritized recommendations, linking claims to operational signals and failure semantics.

\subsection{Case Study B: Enterprise-Scale Deployment}

The framework was evaluated across $\sim$1{,}500 assets spanning 16 asset classes and 14 sites, reflecting heterogeneous sensor coverage, maintenance histories, and coding practices. Assets were categorized as \emph{Not Enough Data}, \emph{Normal}, or \emph{Needs Attention} based on convergence of operational indicators, maintenance history, and failure semantics.

A substantial portion of assets fell into \emph{Not Enough Data}, demonstrating conservative behavior under sparse evidence. 
This conservative classification behavior is reflected quantitatively in the Mean Insight Count (MIC) trend reported in Section~\ref{sec:evaluation}. Category distributions were stable across sites, suggesting thresholds did not overfit local conditions. For assets flagged \emph{Needs Attention}, insights reflected recurring, interpretable failure mechanisms aligned with asset class and history. The deterministic verification loop prevented unsupported conclusions and ensured risks were tied to explicit evidence.

In practice, the system supported prioritization rather than automation, narrowing the subset of assets requiring review. Practitioner overrides mainly reflected local knowledge or pending operational changes, reinforcing the role of human oversight.

\subsection{Operational Impact and Scalability}

In current enterprise workflows, condition-based analysis typically requires manual retrieval and reconciliation of heterogeneous data across multiple systems. Discussions with practitioners indicate that reviewing a single asset may require approximately 5 minutes in the CMMS to retrieve maintenance records, 10 minutes in a building management system to access sensor and SCADA data, and an additional 10–15 minutes in external analytics or IoT platforms to inspect alerts, trends, and metadata. This results in roughly 20–30 minutes of effort per asset.

In contrast, the proposed framework generates integrated condition insights and recommendations in approximately 15–30 seconds per asset. While human validation remains essential, the automation of evidence integration and structured reasoning substantially reduces analyst overhead.

Under existing processes, only a small fraction of assets can be reviewed regularly, limiting practical adoption of condition-based maintenance at scale. By reducing per-asset analysis time from tens of minutes to seconds, the framework makes broader, systematic coverage operationally feasible.

\section{Evaluation}
\label{sec:evaluation}

We evaluate system behavior using two complementary mechanisms: (1) deterministic validation that assesses rule-following performance , and (2) structured judge audits that quantify how generated insights relate to available evidence and governance constraints.

We tested multiple reasoning backbones (Mistral-Medium, LLaMA-4-Maverick, Granite, and GPT-OSS) and multiple frontier models as evaluators (GPT-4.1 and Claude-4 Sonnet). Unless otherwise stated, results reported below use GPT-OSS as the reasoning backbone and Claude-4 Sonnet as the primary (strict) judge, as this pairing produced representative and conservative audit signals. Metrics are computed per asset snapshot, summarized in this section, and formally defined in Appendix~\ref{sec:appendix-metrics}.

\textbf{Grounding.} Unsupported Claim Rate (UCR) measures the proportion of statements not supported by structured evidence.

\textbf{Diagnostic usefulness.} High Specificity Rate (HSR) measures how often statements reference concrete components, failure modes, or maintenance-relevant details.

\textbf{Reasoning stability.} Contradiction Rate (CR) and Redundancy Rate (RR) capture internal inconsistencies and repeated issue descriptions.

\textbf{Coverage and caution.} Mean Insight Count (MIC) reflects output verbosity and responsiveness to available evidence.

\textbf{Rule compliance.} Condition Agreement Rate (CAR) measures alignment between the model’s overall condition classification and a deterministic rule-based condition-selection specification.

\subsection{Effect of Evidence Enrichment and Prompt Constraints}

Table~\ref{tab:judge} isolates two architectural factors: evidence scope (work orders + asset profile vs.\ full structured evidence) and prompt configuration (Naive vs.\ Constrained). The Naive prompt (see Appendix~\ref{appendix:naive_prompt}) requests an unconstrained summary of asset condition, whereas the Constrained prompt explicitly encodes rule-aligned condition criteria and failure-semantic constraints consistent with the system’s deterministic governance logic.

\textbf{Effect of evidence enrichment.}  
Under naive prompting, expanding evidence to include meter abstractions and structured failure-mode knowledge reduces Unsupported Claim Rate (UCR) from 0.007 to 0.003 and slightly increases specificity (HSR: 0.64 → 0.66). This indicates that structured operational and engineering context constrains free-form reasoning and reduces unsupported statements.

\begin{table}[t]
\centering
\small
\begin{tabular}{llcccc}
\toprule
Prompt & Scope & UCR$\downarrow$ & HSR$\uparrow$ & CAR$\uparrow$ & MIC \\
\midrule
Naive & WO & 0.007 & 0.64 & 0.68 & 4.9 \\
Naive & All & 0.003 & 0.66 & 0.70 & 4.9 \\
Constr. & WO & 0.002 & 0.68 & 0.89 & 3.2 \\
Constr. & All & 0.008 & 0.71 & 0.91 & 3.3 \\
\bottomrule
\end{tabular}
\caption{Effect of evidence enrichment and prompt constraints using GPT-OSS evaluated by Claude-4. Prompt constraints substantially improve rule compliance (CAR) and reduce output verbosity (MIC), while structured evidence improves grounding under naive prompting.}
\label{tab:judge}
\end{table}

\textbf{Effect of prompt constraints.}  
The Constrained prompt substantially improves rule compliance across both evidence scopes. Condition Agreement Rate (CAR) increases from 0.68 to 0.89 under limited data and from 0.70 to 0.91 under full evidence. This demonstrates that explicit rule-aligned prompting materially strengthens adherence to deterministic governance logic.

The Constrained prompt also reduces verbosity (MIC: 4.9 → ~3.2), reflecting more selective reasoning. While UCR increases slightly under full evidence with the Constrained prompt (0.003 → 0.008), grounding remains high overall and outputs remain internally stable.

\textbf{Reasoning stability.}  
Contradiction (CR) and redundancy rates (RR) remain near zero across all configurations, indicating that trajectory-controlled reasoning combined with deterministic verification prevents internally conflicting or repetitive issue generation.

\textbf{Variation Across Reasoning Models}
While we evaluated multiple reasoning backbones, observed differences were minor relative to the effects of evidence structure and prompt constraints; detailed per-model results are omitted for brevity.

\textbf{Evaluation Stability Across Judges}
Audits performed with an alternative evaluator (GPT-4.1) yield higher absolute grounding and specificity scores but preserve directional trends. This indicates that observed effects reflect architectural constraints rather than evaluator idiosyncrasies. Deterministic Condition Agreement Rate (CAR), computed independently of LLM judges, provides a stable governance signal across evaluators.

\subsection{Overall Observations}

Results indicate that system behavior is primarily governed by evidence structure and prompt-level constraints rather than backbone model variation. Structured engineering and operational context reduces unsupported reasoning under naive prompting, while constraint-aware prompting dramatically improves rule adherence and reduces verbosity without introducing instability.

The combination of structured evidence construction and deterministic governance therefore provides a controllable reasoning envelope for industrial decision support.

\section{Conclusion and Lessons Learned}
\label{sec:lessons}

Deploying Conditional Insight in enterprise maintenance environments highlights several principles for operationalizing LLM-based reasoning under real-world constraints.

\begin{itemize}
\item \textbf{Conditional explanations improve trust.}
Practitioners preferred structured reasoning framed around “why” and “what next” over raw alerts, aligning outputs with maintenance decision workflows.
\item \textbf{Signal semantics must guide abstraction.}  
Cumulative, incremental, and categorical indicators require distinct summarization strategies; uniform treatment leads to unstable reasoning.
\item \textbf{Failure semantics provide guardrails.}  
FMEA-derived component and mechanism constraints reduce implausible hypotheses and bound the space of admissible explanations.
\item \textbf{LLMs should reason over curated evidence.}  
Operating on structured summaries rather than raw telemetry improves robustness, interpretability, and cross-site consistency.
\item \textbf{Deterministic governance enables production deployment.}  
Separating generative synthesis from rule-based verification ensures reproducibility, auditability, and operational acceptance.
\item \textbf{Prompt design encodes policy.}  
Prompt structure functions as system configuration, requiring iterative refinement aligned with deterministic checks and practitioner feedback.
\end{itemize}

Together, these findings suggest that reliable industrial LLM systems depend more on structured evidence and explicit governance constraints than on backbone model choice alone.



\bibliographystyle{unsrtnat}
\bibliography{custom}  

\clearpage
\appendix

\section{User Scenario Illustration}
\label{sec:appendix_user_scenario}

This section illustrates how the Conditional Insight (CI) framework operates within a maintenance workflow. The scenario demonstrates the interaction between deterministic evidence construction, constrained language reasoning, and rule-based governance.

A maintenance practitioner selects an asset within the CMMS interface to assess its operational condition. Upon selection, the system constructs a structured evidence packet containing asset metadata, work-order history summaries, operational meter abstractions, and FMEA-aligned failure-mode mappings. This evidence is generated deterministically and does not involve language modeling.

The CI agent consumes this structured evidence packet and produces a bounded condition insight consisting of:
\begin{itemize}
    \item an overall condition classification,
    \item evidence-grounded explanatory insights,
    \item prioritized maintenance recommendations.
\end{itemize}

All generated statements are restricted to the provided evidence and admissible failure mechanisms. The overall condition category is subsequently validated against deterministic rule criteria to ensure compliance with operational governance constraints.

Figure~\ref{fig:example_wo_insight} shows an example of the structured condition insight output. The interface separates evidence-based observations from ordered recommendations, explicitly referencing maintenance records and asset attributes. The output format is fixed and enumerated, reflecting the constrained generation and verification design described in the main paper.

The practitioner uses this output to prioritize follow-up actions or further inspection. The system does not trigger automated interventions; it functions strictly as a decision-support layer that narrows attention to evidence-backed hypotheses while preserving human oversight.

\section{Deterministic Meter Abstraction in Conditional Insight}
\label{sec:appendix_meter_abstraction}

\subsection{Upstream Aggregation Model}

Enterprise asset management systems record operational indicators
(“meters”) that summarize physical measurements over time.
Unlike raw high-frequency telemetry (e.g., SCADA streams),
meter readings stored in CMMS platforms typically represent
aggregated or accumulated quantities at curated intervals.

Let $x(t)$ denote an underlying high-frequency signal.
A recorded meter value $v_i$ at time $t_i$ may be expressed as:

\[
v_i = \mathcal{A}\big(x(t)\big),
\qquad
t \in [t_{i-1}, t_i],
\]

where $\mathcal{A}$ denotes an upstream aggregation operator
(e.g., averaging, integration, or accumulation) applied by
the operational system. The Conditional Insight (CI) framework
operates exclusively on these aggregated indicators and does
not ingest raw telemetry.

\subsection{Deterministic Behavioral Abstraction}

Given a meter time series
\[
\{(t_i, v_i)\}_{i=1}^{N},
\]
CI performs deterministic statistical abstraction to convert
meter histories into compact, interpretable behavioral summaries.
These summaries constitute auditable evidence and are incorporated
into the structured \texttt{asset\_facts} packet.

\paragraph{GAUGE Meters.}
For point-in-time measurements (e.g., temperature, pressure),
summary statistics are computed:

\[
\bar{v} = \frac{1}{N}\sum_{i=1}^{N} v_i,
\qquad
s = \sqrt{\frac{1}{N-1}\sum_{i=1}^{N}(v_i - \bar{v})^2}.
\]

Anomalies are identified via Z-score deviations:

\[
z_i = \frac{v_i - \bar{v}}{s},
\qquad
|z_i| > z_{\text{thresh}},
\]

and abrupt changes:

\[
\Delta_i = v_i - v_{i-1}.
\]

Minor oscillations are suppressed using deviation-from-baseline
criteria to ensure only sustained or practically meaningful shifts
are retained.

\paragraph{CONTINUOUS Meters.}
For monotonically increasing totals (e.g., run-hours),
increments are computed:

\[
\Delta_i = v_i - v_{i-1}.
\]

Increment statistics define a normal band:

\[
[\bar{\Delta} - k s_{\Delta},
 \bar{\Delta} + k s_{\Delta}],
\]

where $\bar{\Delta}$ and $s_{\Delta}$ are the mean and
standard deviation of increments.

Event categories include:
\begin{itemize}
    \item Reset events ($\Delta_i < 0$),
    \item Rate anomalies (outside increment band),
    \item Extended flat periods ($|\Delta_i| \le \varepsilon$).
\end{itemize}

\subsection{Structured Evidence Integration}

The output of deterministic abstraction is recorded as
$\texttt{meter\_facts} = \{\text{metadata}, \text{behavioral summaries}, \text{detected events}\}$,
which forms part of the structured \texttt{asset\_facts} evidence packet.

All computations are deterministic and reproducible,
and no raw time-series values are provided to the LLM.
Meter-derived summaries function strictly as bounded,
traceable evidence within the verification-first architecture.

\section{LLM Prompt Design and Guideline}
\label{sec:appendix-schema}

\subsection{Condition Insight Prompt}
\label{appendix:ci_prompt}



The Condition Insight agent’s system prompt defines the agent’s role as an evidence‑grounded maintenance analyst, specifies the reasoning tasks over the structured evidence packet ( \texttt{asset\_facts}), and instructs the model to confine its inferences to the provided evidence. It prescribes strict language and formatting rules (e.g., numeric digits only, expanded terminology, descriptive problem‑code text) and enforces a fixed JSON output schema with the following fields: \emph{Overall Condition}, \emph{Overall Condition Explanation}, \emph{Key insights}, \emph{Recommendations}, and \emph{Overall confidence} (value and reasoning). 

The prompt encodes explicit rule sets for: (i) overall‑condition labels and selection criteria; (ii) key‑insight generation (grounding, component grouping, ordering, and limits); (iii) recommendation formulation (specific component‑level actions using failure‑mode knowledge); and (iv) a deterministic, majority‑rule method for overall confidence. It also provides guidance for interpreting meter patterns to inform insights and recommendations.

At runtime, the prompt incorporates the \texttt{asset\_facts} block which is a structured JSON object produced by the Analytics \& Evidence Construction unit (Fig.~\ref{fig:system_overview}). This evidence packet consolidates information from sensor and meter data, alerts and rule evaluations, work‑order history, and failure‑mode knowledge, as described in Section~\ref{sec:system}, and serves as the sole factual basis for the agent’s reasoning. A simplified JSON schema for this structure is provided in Listing~\ref{lst:assetfacts}. An optional \texttt{{feedback}} block may also be included in the prompt to convey refinement signals during iterative workflows.

\begin{lstlisting}[label={lst:assetfacts},caption={\texttt{asset\_facts} block schema.}]
"asset_facts": {
    "asset_details_facts": {
    asset_number, description, site_ID, priority, status, is_running, failure_code, asset_age_in_years, manufacturer
    },
"workorder_facts": { counts, distributions, preventive_workorders: [...], corrective_and_other_workorders: [...]
    },
"meter_facts": [
      { meter metadata + meter_values[...] }
    ],
"alert_facts": [{ alert metadata... }],
"fmea_facts": [
      { component, mechanism, actions... }
    ],
"health_scores": {
      score_name: { value, range, meaning }
}
\end{lstlisting}

\subsection{Naive Prompt}
\label{appendix:naive_prompt}
In our evaluation results (Section~\ref{sec:evaluation}), we include a naive baseline prompt, shown in Listing~\ref{lst:naive-prompt}, that reflects an unconstrained application of an LLM to the Condition Insight task. This baseline preserves only the JSON output schema required for LLM‑judge evaluation and omits all grounding requirements, verification rules, and domain‑specific constraints used in the deployed CI agent. As a result, the LLM model is free to interpret and reason over the \texttt{asset\_facts} in any manner. This minimal configuration serves as a lower‑bound reference point in our comparative analysis.


\subsection{LLM Judge Evaluation Prompt}
\label{appendix:judge_prompt}
The LLM judge provides a structured, rule‑based assessment of generated insights and recommendations. The LLM evaluation judge is implemented through a two‑part prompt consisting of a system prompt (Listing~\ref{lst:judge-system-prompt}) and a user prompt (Listing~\ref{lst:judge-user-prompt}). The system prompt defines the judge’s role as a strict evaluator, specifying grounding constraints, output‑format requirements, and prohibiting the model from generating any new content beyond the evaluation.

The user prompt provides the full evaluation schema and embeds the following information:
\begin{itemize}
\item \texttt{CONDITION\_INSIGHTS}: the generated condition‑insight summary produced by the Condition Insight agent and subject to evaluation.
\item \texttt{ASSET-FACTS}: the evidence packet provided by the Analytics \& Evidence Construction module to the Condition Insight agent and denoted as \textbf{\{asset\_facts\}}.
\item \texttt{AGENT‑SPECIFICATIONS} (denoted also as \textbf{\{agent\_instructions\}}): the instructions used by the Condition Insight agent's system prompt to generate the condition‑insight summary. 
\end{itemize}

The prompt also enumerates the scoring rules for the evaluation of each insight and recommendation. All per‑item metrics are rated on a 1–3 scale:
\begin{itemize}
\item \textbf{Factuality}: evaluates whether each statement is directly supported by the provided \texttt{asset\_facts}.
\item \textbf{Coherence}: assesses internal logical consistency and the absence of contradictions across items.
\item \textbf{Relevance}: measures the usefulness of each statement for assessing asset condition and maintenance risk.
\item \textbf{Repetitiveness}: quantifies redundancy relative to other insights or recommendations.
\item \textbf{Specificity}: evaluates how concrete, detailed, and data‑anchored each statement is.
\end{itemize}

In addition, the judge assigns two global evaluation metrics:
\begin{itemize}
\item \textbf{Overall‑condition validity} (boolean): determines whether the agent’s overall condition classification is evidence‑supported and compliant with its specification.
\item \textbf{Completeness} (two global scores: one for insights and one for recommendations): measures the number of generated items relative to the expected count.
\end{itemize}
For all scores, the judge is asked to provide a justification that anchors the assigned rating. Taken together, the system and user prompts define the evaluation procedure and promote assessments that are carried out in a controlled and reproducible manner. See the detailed descriptions in Listing~\ref{lst:judge-user-prompt}. Appendix~\ref{sec:appendix-metrics} provides a detailed description on how these scores are aggregated across the multiple experiments to generate the evaluations metrics presented in Section~\ref{sec:evaluation}.


\section{Structured Evaluation Metrics}
\label{sec:appendix-metrics}

We evaluate system behavior using structured judge audits that quantify how generated insights and recommendations relate to available evidence and deterministic rules. Audits are performed using two independent large language models acting strictly as evaluators (Claude‑4 Sonnet and GPT‑4.1), each operating under a fixed evaluation prompt and scoring rubric described in Appendix~\ref{appendix:judge_prompt}. Evaluation metrics are computed directly from the judge outputs, measured per asset snapshot, and then aggregated across evidence scope, reasoning model, and judge model.

\subsection{Grounding Metrics}

\textbf{Unsupported Claim Rate (UCR)} measures the fraction of statements (insights + recommendations) not supported by structured evidence.

\[
\text{UCR} = \frac{\# \text{unsupported statements}}{\# \text{total statements}}
\]

Statements are considered unsupported if no corresponding evidence exists in the structured asset context (e.g., meter summaries, work-order abstractions, or failure mappings). When using the LLM judge (Appendix~\ref{appendix:judge_prompt}), a statement is counted as \emph{unsupported} specifically when the judge assigns it a \textit{Factuality score of 1}.




\subsection{Diagnostic Usefulness}

\textbf{High Specificity Rate (HSR)} measures the proportion of statements that reference concrete components, failure mechanisms, or maintenance-relevant details rather than generic observations.

\[
\text{HSR} = \frac{\# \text{specific, asset-relevant statements}}{\# \text{total statements}}
\]

HSR reflects whether outputs provide actionable, technically meaningful insight. When using the LLM judge (Appendix~\ref{appendix:judge_prompt}), a statement is counted as \emph{specific} when the judge assigns it a \textit{Specificity score of 3}.

\subsection{Reasoning Stability}

\textbf{Contradiction Rate (CR)} measures the fraction of outputs containing internally inconsistent statements.

\[
\text{CR} = \frac{\# \text{outputs with contradictions}}{\# \text{evaluated outputs}}
\]

\textbf{Redundancy Rate (RR)} captures repeated or overlapping issue descriptions within a single output.

\[
\text{RR} = \frac{\# \text{redundant statements}}{\# \text{total statements}}
\]

These metrics assess internal coherence and non-redundant reasoning. When using the LLM judge (Appendix~\ref{appendix:judge_prompt}), an output is counted as containing a contradiction when the judge assigns a \textit{Coherence score of~1}, and a statement is considered redundant when the judge assigns a \textit{Repetitiveness score of~1}.

\subsection{Coverage and Caution}

\textbf{Mean Insight Count (MIC)} is the average number of insights produced per asset snapshot.

\[
\text{MIC} = \frac{\sum \text{insight counts}}{\# \text{asset snapshots}}
\]



It helps characterize how output quantity adapts to evidence availability.

\subsection{Rule Compliance}

\textbf{Condition Agreement Rate (CAR)} is the proportion of assets for which the LLM-assigned condition matches the deterministic rule-based classification. It quantifies rule-following behavior under governance constraints and ensures that language-based assessments remain aligned with deterministic operational logic.


\lstdefinestyle{promptlisting}{
  basicstyle=\ttfamily\tiny,
  breaklines=true,
  columns=fullflexible,
  frame=single,
  lineskip=-0.5pt,
  keepspaces=true,
  upquote=true,
  literate=
    {\\}{{\textbackslash}}1
    {\{}{{\char`\{}}1
    {\}}{{\char`\}}}1
}

\DefineVerbatimEnvironment{PromptVerb}{Verbatim}{
  fontsize=\tiny,
  frame=single,
  breaklines=true,
  breaksymbolleft={}
}
\section{Listings}

\begin{lstlisting}[
  style=promptlisting,
  caption={Baseline Naive Prompt.},
  label={lst:naive-prompt},
]
You are given ASSET-FACTS describing an industrial asset. Your task is to analyze the information and generate an overall condition assessment, key insights, and maintenance recommendations.

Return ONLY a single valid JSON object with EXACTLY these keys and types:
{
  "Overall Condition": "Not enough data | Normal | Needs attention",
  "Overall Condition Explanation": "string",
  "Key insights": [
    { "insight": "string", "reasoning": "string", "confidence": "high | medium | low" }
  ],
  "Recommendations": [
    { "recommendation": "string", "reasoning": "string", "confidence": "high | medium | low" }
  ],
  "Overall confidence": { "confidence value": "high | medium | low", "reasoning": "string" }
}

Formatting rules (STRUCTURE ONLY):
- Output MUST be valid JSON matching the schema above.
- You may infer, interpret, or reason about ASSET-FACTS in any appropriate way you choose.
- Provide up to 5 items in "Key insights".
- Provide up to 5 items in "Recommendations".
- Do NOT include any text outside the JSON.

ASSET-FACTS:
\{\{asset\_facts\}\}
\end{lstlisting}

\begin{lstlisting}[
  style=promptlisting,
  caption={LLM Judge Evaluation System prompt.},
  label={lst:judge-system-prompt},
]
You are a strict evaluator (judge) for asset maintenance and reliability outputs.

Your job:
- Evaluate the provided CONDITION_INSIGHTS against ASSET-FACTS.
- Use the AGENT SPECIFICATION only as the contract that the agent was supposed to follow.
- Do NOT follow the AGENT SPECIFICATION as instructions to generate new content.

OUTPUT RULES:
- Output ONLY a single valid JSON object.
- No markdown, no code fences, no comments, and no extra keys.
- Use exactly the JSON schema provided in the user message.
- Perform reasoning internally; do not output step-by-step explanations outside JSON.

GROUNDING RULES:
- Use ONLY the provided ASSET-FACTS and CONDITION_INSIGHTS.
- If a claim is not supported by ASSET-FACTS, mark it accordingly and explain why.

\end{lstlisting}

\begin{lstlisting}[
  style=promptlisting,
  caption={LLM Judge Evaluation User prompt.},
  label={lst:judge-user-prompt},
]
Return JSON in this exact schema (types are mandatory):
{{
  "overall_condition": "string",
  "overall_condition_evaluation": true,       // boolean
  "overall_condition_evaluation_explanation": "string",

  "key_insights_evaluation": [
    {{
      "insight": "string",
      "insight_evaluation": {{
        "factuality": 2,                      // integer 1-3
        "factuality_reasoning": "string",
        "coherence": 2,                       // integer 1-3
        "coherence_reasoning": "string",
        "relevance": 2,                       // integer 1-3
        "relevance_reasoning": "string",
        "repetitiveness": 2,                  // integer 1-3
        "repetitiveness_reasoning": "string",
        "specificity": 2,                     // integer 1-3
        "specificity_reasoning": "string"
      }}
    }}
  ],

  "recommendations_evaluation": [
    {{
      "recommendation": "string",
      "recommendation_evaluation": {{
        "factuality": 2,                      // integer 1-3
        "factuality_reasoning": "string",
        "coherence": 2,                       // integer 1-3
        "coherence_reasoning": "string",
        "relevance": 2,                       // integer 1-3
        "relevance_reasoning": "string",
        "repetitiveness": 2,                  // integer 1-3
        "repetitiveness_reasoning": "string",
        "specificity": 2,                     // integer 1-3
        "specificity_reasoning": "string"
      }}
    }}
  ],

  "recommendations_completeness": 2,           // integer 1-3
  "recommendations_completeness_reasoning": "string",

  "insights_completeness": 2,                  // integer 1-3
  "insights_completeness_reasoning": "string"
}}

RATING SCALE (applies to all 1-3 fields):
- 1 = not satisfied
- 2 = partially satisfied
- 3 = fully satisfied

RULES FOR OVERALL CONDITION EVALUATION:
- "overall_condition" MUST be copied verbatim from the agent's CONDITION_INSIGHTS output.
- "overall_condition_evaluation" MUST be true if and only if BOTH are satisfied:
  (1) The agent's overall condition is supported by ASSET-FACTS, AND
  (2) The agent's overall condition follows the "Rules for Overall Condition" in the AGENT SPECIFICATION.
- Otherwise, "overall_condition_evaluation" MUST be false.
- The explanation must cite specific ASSET-FACTS evidence (e.g., corrective work orders count, overdue work orders, anomalies, alert totals, health score ranges).

RULES FOR FACTUALITY (applies to each insight or recommendation individually):
Rate factuality on a 1-3 scale based ONLY on how well the statement is grounded in ASSET-FACTS.
- Score 1 (not grounded): The statement is unsupported or contradicted by ASSET-FACTS.
- Score 2 (partially grounded): Some parts are supported by ASSET-FACTS, but other parts extend beyond the data, are ambiguous, or weakly justified.
- Score 3 (fully grounded): The statement is completely and directly supported by ASSET-FACTS.
For "factuality_reasoning": Explain which ASSET-FACTS elements support or fail to support the statement.

RULES FOR COHERENCE (applies to each insight or recommendation individually):
Rate coherence on a 1-3 scale based on whether the statement is internally consistent and non-contradictory with other insights or recommendations.
- Score 1 (contradicting): The statement contradicts itself or contradicts other insights or recommendations.
- Score 2 (partially coherent): Mostly consistent, but contains unclear logic, minor contradictions, or ambiguous connections.
- Score 3 (fully coherent): Completely consistent with itself and with all other insights and recommendations; no contradictions or logical conflicts.
For "coherence_reasoning": Explain how the score was derived, citing specific inconsistencies or confirming consistency.

RULES FOR RELEVANCE (applies to each insight or recommendation individually):
Rate relevance on a 1-3 scale based on how meaningful and useful the statement is for asset maintenance and evaluating Overall Condition.
- Score 1 (irrelevant): The statement does not meaningfully affect asset condition, risk, failures, or maintenance decisions; may be trivial or routine (e.g., stating something works fine without impact).
- Score 2 (somewhat relevant): The statement relates to meaningful topics but is generic, incomplete, or weakly connected to maintenance priorities or the rules.
- Score 3 (highly relevant): The statement directly addresses asset risks, failures, condition drivers, or actionable maintenance considerations, and strongly aligns with the rules.
For "relevance_reasoning": Explain why the statement is or is not meaningful for maintenance decisions and Overall Condition.

RULES FOR REPETITIVENESS (applies to each insight or recommendation individually):
Rate repetitiveness on a 1-3 scale based on how much the statement repeats content already provided in other insights or recommendations.
- Score 1 (very repetitive): The statement repeats the same idea, issue, component, evidence, or conclusion as another item with little or no new information.
- Score 2 (somewhat repetitive): The statement overlaps partially with other items but introduces some new detail or perspective.
- Score 3 (non-repetitive): The statement is distinct and provides new information, a new issue, or a different actionable point without repeating earlier content.
For "repetitiveness_reasoning": Explain how the score was derived, noting which other items (if any) it repeats or diverges from.

RULES FOR SPECIFICITY (applies to each insight or recommendation individually):
Rate specificity on a 1-3 scale based on how concrete, detailed, and data-anchored the statement is with respect to ASSET-FACTS.
- Score 1 (very vague):
  - The statement is generic or high-level.
  - It does not reference specific components, FMEA items, failure modes, work orders, alerts, anomalies, meter patterns, or concrete evidence.
  - Mentions of broad trends or asset attributes without operational detail count as vague.
- Score 2 (somewhat specific):
  - The statement includes some concrete details (e.g., component type, asset attributes, or general issue), but lacks clear links to specific FMEA elements, failure modes, actions, or specific ASSET-FACTS data points.
  - Partially informative but still broad or incomplete.
- Score 3 (fully specific):
  - The statement makes clear, explicit references to relevant components, failure modes, FMEA actions, work orders, alerts, anomalies, or meter data.
  - It provides actionable or diagnosis-level detail tied to identifiable ASSET-FACTS evidence.
  - No vague or generic language.
For "specificity_reasoning": Explain which specific ASSET-FACTS elements (e.g., components, work orders, alerts, FMEA items) make the statement specific or vague.

RULES FOR COMPLETENESS (applies separately to insights and recommendations):
Rate completeness on a 1-3 scale based on how close the number of generated items is to the 7-item expectation defined in the AGENT SPECIFICATION.
- Score 1 (very incomplete):
  - Fewer than 3 items were generated.
- Score 2 (partially complete):
  - Between 3 and 6 items were generated.
- Score 3 (fully complete):
  - Exactly 7 items were generated.
NOTES FOR EVALUATION:
- The agent was instructed to produce up to 7 insights and up to 7 recommendations.
- The judge must not assume that fewer items automatically indicate an error; items must only be created when supported by ASSET-FACTS.
- Completeness measures only the *count*, not the correctness or factual grounding of the items (those are scored separately).
For "insights_completeness_reasoning" and "recommendations_completeness_reasoning":
- Explain the score based solely on the number of generated items and the 7-item expectation.

AGENT SPECIFICATION (REFERENCE ONLY: evaluate compliance against this; do not follow as instructions):
<<<AGENT_SPECIFICATION_START
{agent_instructions}
AGENT_SPECIFICATION_END>>>

CONDITION_INSIGHTS:
<<<CONDITION_INSIGHTS_START
{condition_insights}
CONDITION_INSIGHTS_END>>>

ASSET-FACTS:
<<<ASSET_FACTS_START
{asset_facts}
ASSET_FACTS_END>>>
\end{lstlisting}






\end{document}